\documentclass[9pt,twocolumn,twoside]{pnas-new}

\templatetype{pnasresearcharticle}
\usepackage{eqnarray,amsmath,multirow}

\title{Automatic classification of multiple catheters in neonatal radiographs with deep learning}

\author[a,1,$\dag$]{Robert D. E. Henderson MBA, PhD}
\author[a,$\dag$]{Xin Yi PhD} 
\author[a]{Scott J. Adams MD}
\author[a]{Paul Babyn MDCM}

\affil[a]{Department of Medical Imaging, College of Medicine, University of Saskatchewan, Canada}
\affil[$\dag$]{Equal contribution to this work}

\leadauthor{Henderson} 

\significancestatement{A convolutional neural network was developed to detect and simultaneously classify multiple catheters on neonatal radiographs using a multi-label approach, achieving high accuracy for four common catheter types.}

\authordeclaration{The authors declare no conflicts of interest.}
\correspondingauthor{\textsuperscript{1}To whom correspondence should be addressed. E-mail: robert.henderson@usask.ca}


\begin{abstract}
We develop and evaluate a deep learning algorithm to classify multiple catheters on neonatal chest and abdominal radiographs.
A convolutional neural network (CNN) was trained using a dataset of 777 neonatal chest and abdominal radiographs,
with a split of 81\%-9\%-10\% for training-validation-testing, respectively. We employed ResNet-50 (a CNN), pre-trained on
ImageNet. Ground truth labelling was limited to tagging each image to indicate the presence or absence of endotracheal tubes (ETTs),
nasogastric tubes (NGTs), and umbilical arterial and venous catheters (UACs, UVCs). The data set included 561 images containing 2 or
more catheters, 167 images with only one, and 49 with none. Performance was measured with average precision (AP), calculated from
the area under the precision-recall curve. On our test data, the algorithm achieved an overall AP (95\% confidence interval) of 0.977
(0.679–0.999) for NGTs, 0.989 (0.751–1.000) for ETTs, 0.979 (0.873–0.997) for UACs, and 0.937 (0.785–0.984) for UVCs.
Performance was similar for the set of 58 test images consisting of 2 or more catheters, with an AP of 0.975 (0.255–1.000) for 
NGTs, 0.997 (0.009–1.000) for ETTs, 0.981 (0.797–0.998) for UACs, and 0.937 (0.689–0.990) for UVCs. Our network thus achieves
strong performance in the simultaneous detection of these four catheter types. Radiologists may use such an algorithm as a time-saving
mechanism to automate reporting of catheters on radiographs.
\end{abstract}

\dates{This manuscript was compiled on \today.}
\doi{\url{www.arxiv.org/content/doi}}

\begin{document}

\maketitle
\thispagestyle{firststyle}
\ifthenelse{\boolean{shortarticle}}{\ifthenelse{\boolean{singlecolumn}}{\abscontentformatted}{\abscontent}}{}

\section*{Introduction}
Assessment of catheters and tubes on radiographs is a common, important, and occasionally meticulous task for radiologists \cite{Schmidt2008,Remerand2007,Thomas1998,Godoy2012}. Neonatal chest and abdomen radiographs amplify these characteristics \cite{MacDonald2013}: i) many neonates require such radiographs to verify catheter placement; ii) catheter placement must be urgently assessed; and iii) some types of catheters in neonates have unique routes through umbilical vessels which can be more challenging to assess \cite{Green1998}. Some of the most common catheters and tubes (henceforth, simply, catheters) include nasogastric tubes (NGTs) to support feeding, endotracheal tubes (ETTs) to support breathing, umbilical arterial catheters (UACs) for blood pressure monitoring, and umbilical venous catheters (UVC) for central venous access \cite{Concepcion2017}. Each of these catheters require correct detection, identification, and analysis of placement with respect to anatomy to determine if they are in the intended position for their purpose.

Computer-aided methods may be used in the future to augment the workflow of radiologists by, for example, flagging certain cases for urgent review, or recording the presence and placement of catheters to expedite reporting. Thus far, studies have been conducted mostly of single catheter types or with rules-based approaches with respect to catheter appearance or location \cite{Keller2007a,Huo2007,Brunelli2009a,Duda1972}. Artificial intelligence, and more specifically deep learning approaches, have become the new standard for catheter assessment \cite{Lakhani2017,Singh2019,Mercan2014,Lee2018,Yi2019,Frid-Adar2019}. However, few aim to analyse multiple different types of catheters \cite{Subramanian2019}.

An extensive review of the current state-of-the-art for artificial intelligence applied to catheter assessment was recently published \cite{Yi2020a}, and we refer the reader there for more detail. In brief, there are four main questions related to assessment of catheters on radiographs: i) is there a catheter present? ii) which type of catheter is it? iii) what is the course of the catheter? and iv) where is the tip of the catheter? When combined, these lead to a fifth and final question: is the catheter in a satisfactory position?

The present work focusses on answering questions (i) and (ii) together to detect the presence of one or more catheters and identify their type (NGT, ETT, UAC, or UVC). Analysis of only the presence of catheters as in the first question (i) has been reported before for single catheters \cite{Huo2007,Lakhani2017,Sheng2009,Ramakrishna2011,Chen2016,Keller2007a,Kao2015} and generalised to any type by Yi et al. \cite{Yi2019}. However, this approach is either limited to one catheter, is blind to catheter type, or is constrained by user-defined features. Here, we employ a deep convolutional neural network (CNN) that makes no underlying assumptions about what is important in the detection and classification of the four previously mentioned catheters. Moreover, our model incorporates multiple catheter types into a single network.

In the following, we first present the details of our CNN architecture and dataset construction. We then describe the metrics we use to analyse our model performance and show our results for each catheter type. Importantly, we also analyse performance based upon the number of catheters in each image (e.g., those with none, one, or more than one). These results are then discussed in the context of previously reported work, both in the realm of catheter classification, and compared to classification results of other radiographic abnormalities. Thus, here we specifically address the issue of automatic detection and classification when multiple catheter types are present. This work furthers research toward a complete model that can automatically – and with high accuracy – determine if a catheter is correctly placed. A future tool such as this might be used to improve efficiency and accuracy of radiologist reporting.

\section*{Methods}
\subsection*{Dataset and labelling}
A dataset of 777 de-identified neonatal antero-posterior view chest/abdominal radiographs was compiled from our local institutional Picture Archiving and Communication System (PACS). Cases were limited to neonatal studies, of variable lung and abdominal pathology, chosen consecutively from 2014-2015 at Royal University Hospital, Saskatoon, Saskatchewan, Canada. Each image was obtained within the local NICU using standard protocols. This work was considered exempt from full review by the University of Saskatchewan Research Ethics Board. Minor subsets of this dataset have been published in other works as follows: i) Yi et al., 2019 \cite{Yi2019}, where 50 of these cases were used as a test dataset on a model that predicted the presence of lines and tubes on neonatal radiographs from synthetic training data, and ii) Yi et al., 2020 \cite{Yi2020a}, a review article on computer-aided assessment of catheters and tubes where 100 cases were annotated for public use. However, these studies differ from the present work as neither focused on classification of catheter types, nor have the data been previously used to train a CNN.

Each radiograph was assigned a label to indicate only the presence or absence of each type of catheter in our consideration: NGT, ETT, UAC, and UVC. To facilitate this process, we used a custom-built Python script to display images sequentially and record identified catheters into a datafile. Labels were first recorded by a medical student, and subsequently reviewed by a resident and an attending paediatric radiologist. A summary of the dataset is shown in Table\,\ref{tab:data}. In brief, we split the dataset at random into approximate 81\%-9\%-10\% percentages for training-validation-testing.

\begin{table*}[ht]
\centering
\caption{Image counts for each subset of the data, by number and type of label (catheter type).}
\begin{tabular}{llllllllllll}
           & Images   & \multicolumn{5}{l}{Number of images   containing 0 – 4  catheters/labels (\%):} & \multicolumn{5}{l}{Number of images   containing each catheter:} \\ \cmidrule(lr){3-7} \cmidrule(lr){8-12}
           & (\%)     & 0            & 1              & 2              & 3              & 4          & $~~~~$  & NGT            & ETT            & UAC            & UVC           \\ \midrule
Training   & 629 (81) & 38 (6)       & 138 (22)       & 198 (31)       & 126 (20)       & 129 (21)     && 490            & 367            & 219            & 352           \\
Validation & 70 (9)   & 4 (6)        & 16 (23)        & 22 (31)        & 13 (19)        & 15 (21)       && 56             & 45             & 21             & 37            \\
Testing    & 78 (10)  & 7 (9)        & 13 (17)        & 24 (31)        & 19 (24)        & 15 (19)       && 59             & 47             & 27             & 45            \\ \midrule
Total      & 777      & 49           & 167            & 244            & 158            & 159           && 605            & 459            & 267            & 434  \\ \bottomrule
\end{tabular}
\label{tab:data}
\end{table*}

\subsection*{CNN architecture and treatment}
\begin{figure*}[tbhp]
\centering
\includegraphics[width=6.8in]{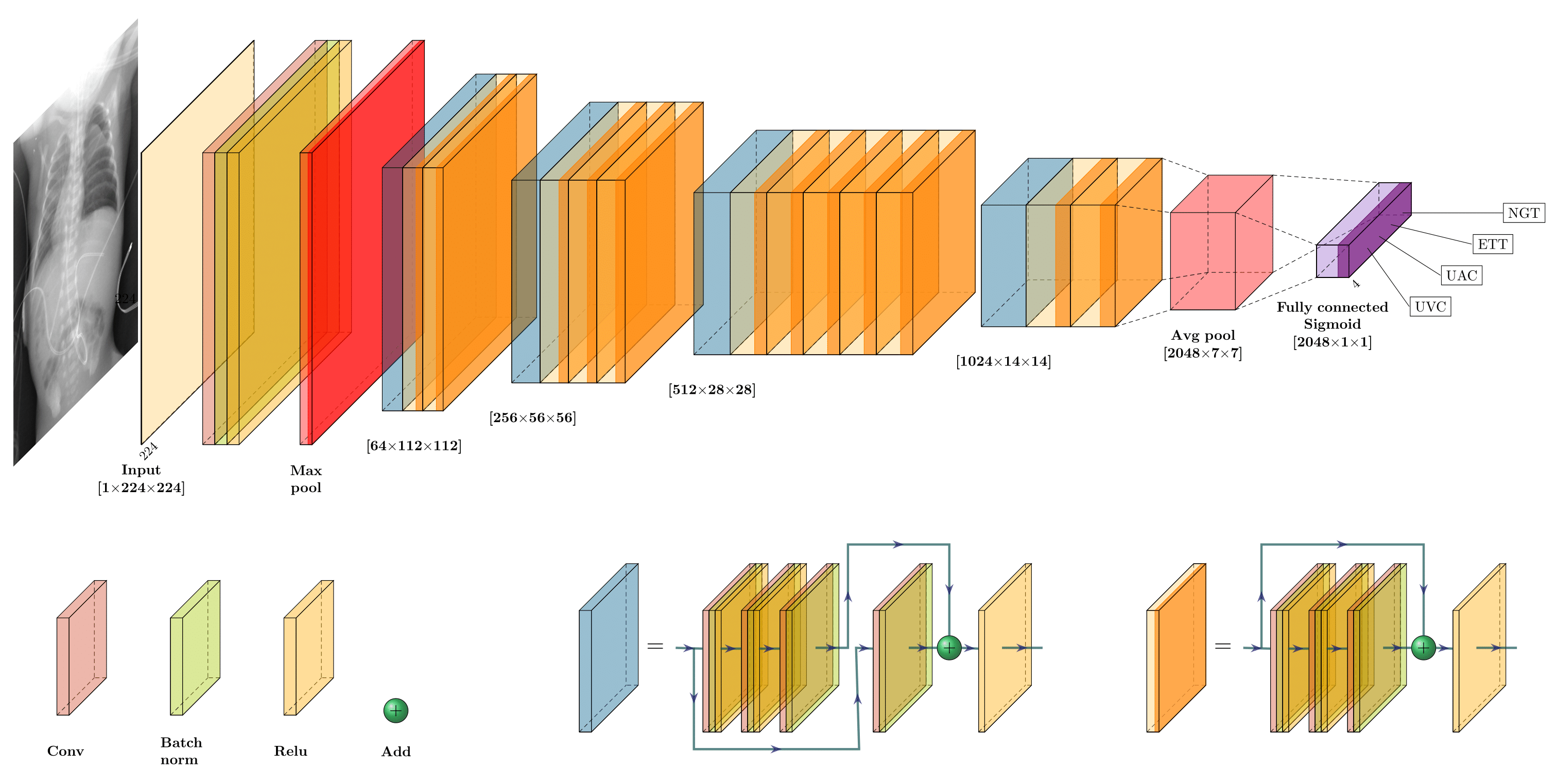}
\caption{A graphical representation of our CNN architecture based upon ResNet-50 \cite{He2016a}. The main structure above starts at the left with the image as input, working to the right as each layer acts as input to the next; in this depiction each block represents the output feature map of the underlying layer operation. A basic building block of the Resnet-50 architecture is the conv-bn-relu sequence consisting of a convolutional layer (Conv), a batch normalisation (Batch norm, or bn), and a rectified linear unit layer (ReLU). Detailed structure is shown in the lower half of the figure for compactness, wherein some layers are either added together or rerouted. As the network progresses, the convolution reduces the (x,y) dimensions while increasing the z (not to scale).}\label{fig:model}
\end{figure*}

Our base CNN was ResNet-50 \cite{He2016a} pretrained on ImageNet and fine-tuned on our dataset. The images were resized to 224x244 pixels to be compatible with the input size of ResNet-50, and we employed horizontal flipping for data augmentation. The last fully connected layer of ResNet-50, of which there were 1000 output nodes, was replaced with another fully connected layer consisting of 4 output nodes to represent the 4 labels for catheter types (NGT, ETT, UAC, UVC); each output node yields a probability for the presence or absence of that catheter. See Fig.\,\ref{fig:model} for a graphical representation of our CNN setup. This setup formed the basis for our multi-label network. Training was optimized using stochastic gradient descent with momentum 0.9. A learning rate of 0.001 was chosen and halved after each set of 20 epochs. We used a batch size of 16 images, a total of 50 epochs, and used cross entropy for the loss function. We chose the optimal point at which to stop training based upon the lowest point of the validation loss. In addition to the above, individual analogous networks were trained using the same base architecture, but each with a single output node, one for each catheter type; these are denoted “single-label” networks and are retained for comparison to the multi-label network.

\subsection*{Performance assessment}
We measured performance of the network on our test data with: i) average precision (AP), computed as the area under the precision-recall curve for each of our labels; ii) sensitivity and specificity; and iii) Hamming loss. Hamming loss is defined as [1 – Accuracy], or simply the fraction of incorrectly predicted labels to the total number of labels; this measure is helpful for cases of imbalanced multi-label classification as in our dataset \cite{Tsoumakas2007}. Our threshold probabilities, which indicate the predicted presence (at or above threshold) or absence (below threshold) of each catheter, were derived from the validation dataset. This avoids overestimating the performance of the network by tailoring the thresholds to the held-out test dataset. Thresholds for each label were selected such that the quantity (Sensitivity + Specificity) was maximised \cite{Zou2013} (see also Fig.\,\ref{fig:prvalsupp} in the Supplementary Data). With this method, the thresholds were chosen to be: 0.8 (NGT), 0.2 (ETT), 0.8 (UAC), and 0.75 (UVC). The precision-recall curve is constructed from a series of values of precision (the fraction of true positive cases in all positive tests) and recall (sensitivity) for various settings of the threshold probability, and thus average precision is a measure independent of the chosen thresholds.

Class activation maps (CAMs) are a useful tool to visualize which regions in each image influence the prediction for each output label \cite{Singh2019,Zhou2016}. In our case we are not classifying images as belonging to a single catheter; thus, to avoid confusion we will refer to these as label activation maps (LAMs) in this work, and compute them for each label. 

For later analysis (not for training), we placed each test image into a set based upon the number of catheters present, including a set of all test images that contain exactly 1 catheter, another set with exactly 2 catheters, and so on. In this work, “multiple catheters” refers to the set of images containing two or more catheters. Thus, we refer to “labels” when discussing individual catheter types and output from the deep learning algorithm, and used these additional sets to interpret performance with respect to the number of catheters present in the images. 

The additional metrics of sensitivity, specificity, and Hamming loss permit an evaluation of performance in those instances where AP is undefined, such as in the set of images with no catheters where there is an absence of true positive cases. Confidence intervals were calculated by treating each metric as an estimate of the probability parameter of the associated binomial random variable. For instance, sensitivity is the probability of labelling a positive case as positive; analogous arguments can be made for the other metrics. That is, metrics such as sensitivity must vary only between 0 and 1, and are de facto probabilities. Since it is established in the literature \cite{Sternberg2001} that it is inappropriate to use a symmetric confidence interval when the point estimate (i.e. the sensitivity value calculated based on the sample) is close to either 0 or 1, a logit transformation was used to calculate the confidence bounds \cite{Sternberg2001,Boyd2013}.

\section*{Results}
\subsection*{Network performance}
As introduced above, Table\,\ref{tab:data} describes our dataset split by type of catheter and number of images containing 0, 1, …, or 4 catheters each. NGTs are the most common (present on 605 of 777 radiographs), and UACs the least common (present on 267 of 777 radiographs). Table\,\ref{tab:results} summarizes the performance for our network, with additional detail provided in Supplementary Data (Table\,\ref{tab:resultssupp}). It is noted from Table 2 that performance, generally, is high for both the multi-label and single-label networks. Performance on the set of test images that contained multiple catheters (AP 0.937 – 0.997; sensitivity 0.741 – 0.950; specificity 0.667 – 1.00; ranges across all catheter types) is similar to performance on the set of all test images (AP 0.937 – 0.989; sensitivity 0.741 – 0.956; specificity 0.788 – 1.00). Of the 13 single-catheter images across all 4 catheter label predictions (for a total of 52 predictions) only two false positives were identified (see Table\,\ref{tab:counts}). AP is the lowest for detection and classification of UVCs, while ETTs perform most poorly in terms of specificity; though due to the sample size these are not statistically significant. Generally, AP is greater than 0.9 for the overall multi-label and single-label networks across all catheter types.

\begin{table*}[ht]
\centering
\caption{Results overview of 78 test cases for each catheter type from our model.}
\begin{tabular}{lllll}
                                   &      & \multicolumn{2}{l}{Multi-label network} & Single-label network$^c$ \\
\multicolumn{2}{l}{} & \multicolumn{2}{l}{Number of labels per  image:} & \\
                                   &      & \textgreater{}1$^a$                                          & 0 – 4$^b$                                                    & 0 – 4$^b$                \\ \cmidrule(lr){3-4} \cmidrule(lr){5-5}
\multirow{4}{*}{Average Precision} & NGT  & 0.975 (0.255   – 1.000)                                   & 0.977   (0.679 – 0.999)                                   & 0.976 (0.683 – 0.999) \\
                                   & ETT  & 0.997   (0.009 – 1.000)                                   & 0.989 (0.751   – 1.000)                                   & 0.982 (0.794 – 0.999) \\
                                   & UAC  & 0.981   (0.797 – 0.998)                                   & 0.979 (0.873   – 0.997)                                   & 0.979 (0.873 – 0.997) \\
                                   & UVC  & 0.937   (0.689 – 0.990)                                   & 0.937 (0.785   – 0.984)                                   & 0.917 (0.763 – 0.975) \\ \midrule
\multirow{5}{*}{Sensitivity}       & NGT  & 0.941   (0.833 - 0.981)                                   & 0.949   (0.854 - 0.984)                                   & 0.947 (0.706 - 0.993) \\
                                   & ETT  & 0.915 (0.794   - 0.968)                                   & 0.915   (0.794 - 0.968)                                   & 0.867 (0.595 - 0.966) \\
                                   & UAC  & 0.741   (0.547 - 0.871)                                   & 0.741   (0.547 - 0.871)                                   & 0.500 (0.123 - 0.877) \\
                                   & UVC  & 0.950   (0.821 - 0.987)                                   & 0.956   (0.839 - 0.989)                                   & 0.911 (0.786 – 0.966) \\
                                   & All$^d$ & 0.903   (0.848 - 0.940)                                   & 0.910   (0.858 - 0.944)                                   & 0.896 (0.773 - 0.956) \\ \midrule
\multirow{5}{*}{Specificity}       & NGT  & 0.857   (0.419 – 0.980)                                   & 0.895 (0.663   – 0.974)                                   & 0.842 (0.608 – 0.948) \\
                                   & ETT  & 1 ( - )                                                   & 0.967 (   0.804 – 0.995)                                  & 0.710 (0.530 – 0.841) \\
                                   & UAC  & 1 ( - )                                                   & 1 ( - )                                                   & 0.980 (0.874 – 0.997) \\
                                   & UVC  & 0.667   (0.429 – 0.842)                                   & 0.788 (0.617   – 0.895)                                   & 0.788 (0.617 – 0.895) \\
                                   & All$^d$ & 0.896   (0.797 – 0.949)                                   & 0.925   (0.867 – 0.959)                                   & 0.851 (0.780 – 0.902) \\ \midrule
\multirow{5}{*}{Hamming loss$^e$}     & NGT  & 0.035   (0.009 – 0.128)                                   & 0.039   (0.012 – 0.113)                                   & 0.051 (0.019 – 0.129) \\
                                   & ETT  & 0.138   (0.071 – 0.252)                                   & 0.115   (0.061 – 0.207)                                   & 0.128 (0.070 – 0.222) \\
                                   & UAC  & 0.069   (0.026 – 0.170)                                   & 0.051 (0.019   – 0.129)                                   & 0.064 (0.027 – 0.145) \\
                                   & UVC  & 0.103   (0.047 – 0.212)                                   & 0.090   (0.043 – 0.176)                                   & 0.141 (0.080 – 0.237) \\
                                   & All$^d$ & 0.086 (0.056 – 0.130)                                     & 0.074 (0.049 – 0.108)                                     & 0.097 (0.068 – 0.134) \\ \bottomrule
\end{tabular}
\label{tab:results}
\addtabletext{$~$\\
Confidence intervals are calculated at 95\% as described in the text.\\
$^a$ Combined results for those images containing multiple (2 or more) catheters.\\
$^b$ Combined results for all images (containing 0 to 4 catheter types).\\
$^c$ Single-label results are for individual binary classification networks for each type of catheter.\\
$^d$ When computing these measures for all catheter types, there is some dependence among the data points since most of the images contain multiple catheters which may influence each other. Thus, this metric is not robust.\\
$^e$ Note that, unlike the three previous measures, values closer to zero represent better performance. See the text for details.\\}
\end{table*}

\begin{figure}[tbhp]
\centering
\includegraphics[width=\linewidth]{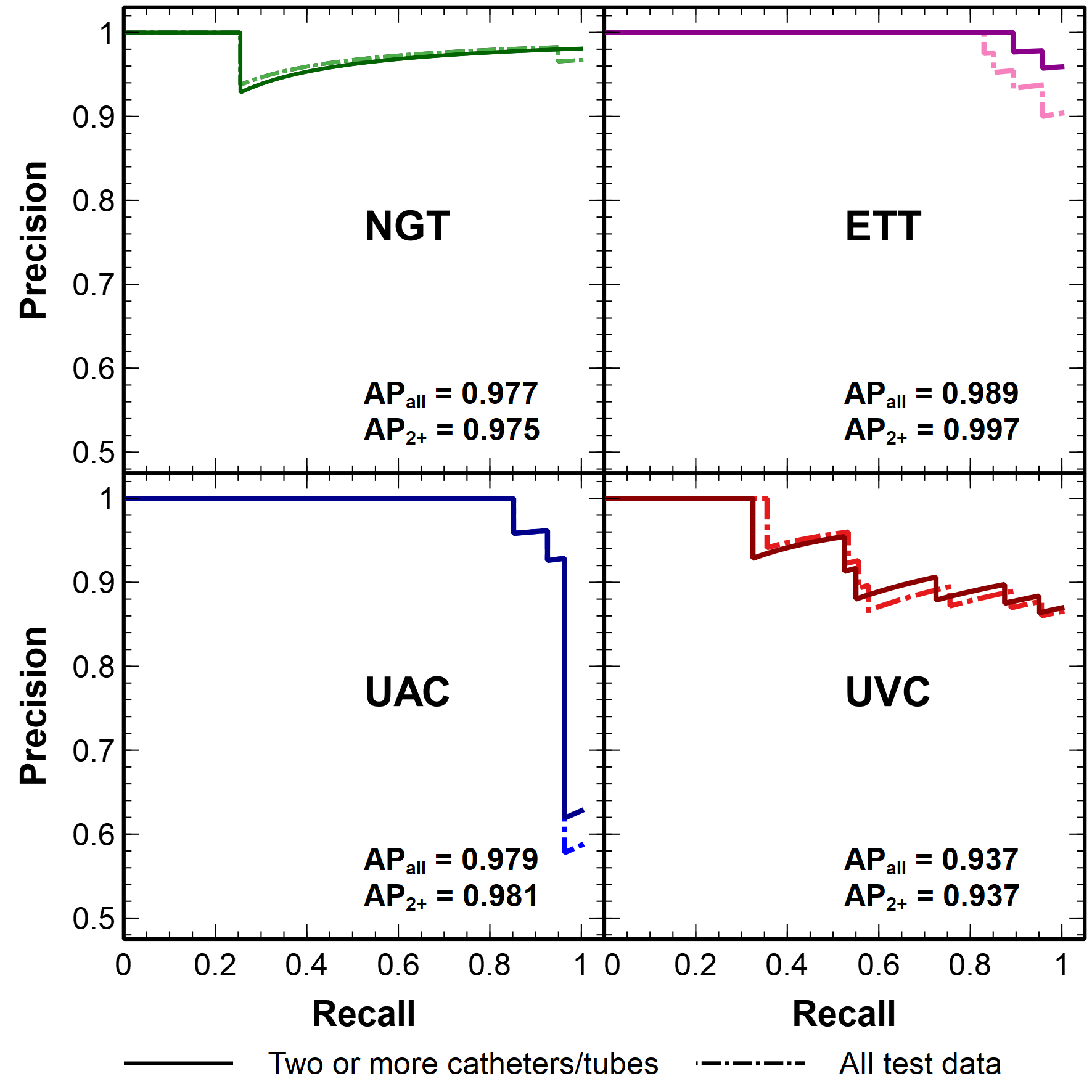}
\caption{Precision-Recall curves (from the test dataset) for our 4 catheters of interest. We show results for the entire test dataset (dash-dot curves) and for only those test images containing multiple (2+) catheters (solid curves). Average precision is indicated for each curve within the plots. AP – average precision; NGT – nasogastric tube; ETT – endotracheal tube; UAC – umbilical arterial catheter; UVC – umbilical venous catheter.}
\label{fig:pr}
\end{figure}

\begin{figure}[tbhp]
\centering
\includegraphics[width=\linewidth]{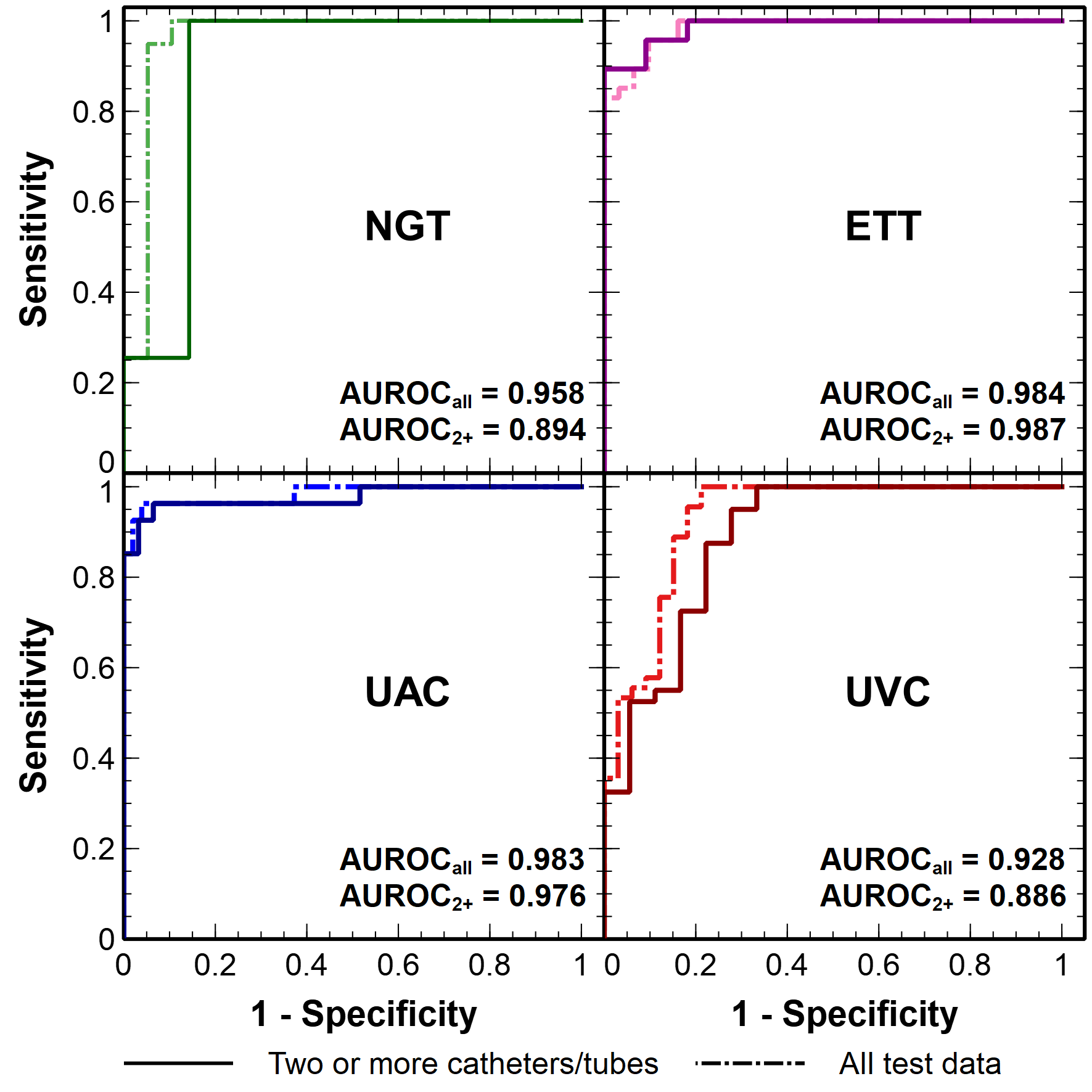}
\caption{Receiver operating characteristic curves (from the test dataset) for our 4 catheters of interest; lines as in Fig. 1. AUROC – area under the receiver operating characteristic curve; NGT – nasogastric tube; ETT – endotracheal tube; UAC – umbilical arterial catheter; UVC – umbilical venous catheter.}
\label{fig:roc}
\end{figure}

In Figs.\,\ref{fig:pr} and \ref{fig:roc} we show the precision-recall curves and receiver operating characteristic (ROC) curves, respectively, for each catheter type, both for the entire test dataset and those images containing multiple catheters (calculations of the area under the ROC curve, AUROC, are included in Table\,\ref{tab:resultssupp}. In each case the curves are favourable, and comparable for the two subsets containing multiple catheters and the entire dataset. As is also shown by the overall computed measures, UVCs have the least optimal precision-recall curves (see AP values above and in Tables\,\ref{tab:results} and \ref{tab:resultssupp}).

\subsection*{Visualising performance}
\begin{SCfigure*}[\sidecaptionrelwidth][ht]
\centering
\includegraphics[width=14.5cm]{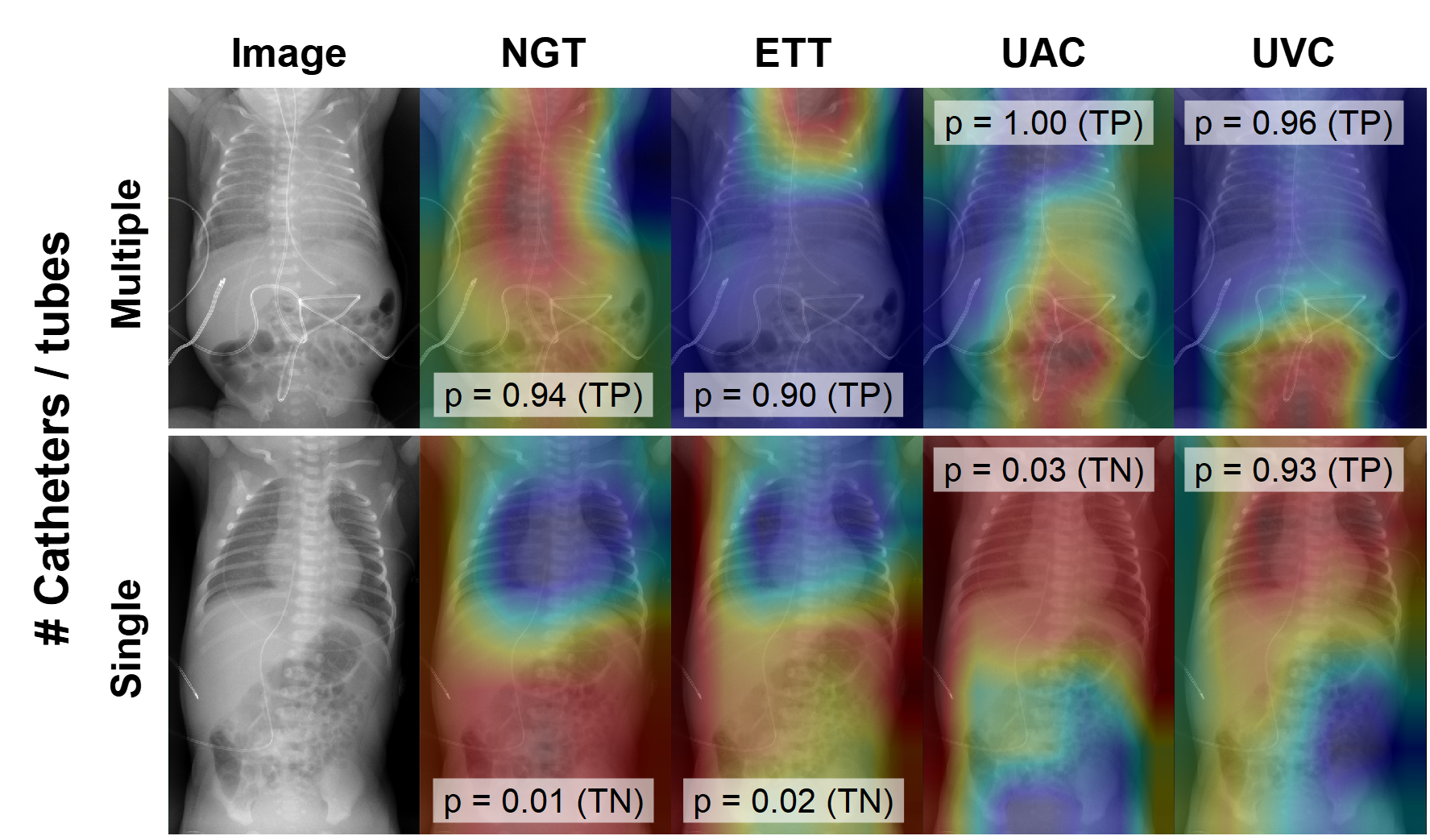}
\caption{Examples of CAMs for successful catheter classification on images containing single or multiple catheters of interest (bottom and top rows, respectively). Red indicates regions of high influence on the final classification result; blue indicates regions of lower influence. Catheters expected to be present show probability p > 0.5, while those considered to be absent have p < 0.5.}
\label{fig:correct}
\end{SCfigure*}

Example label activation maps of successful prediction for a single-catheter and (all) 4-catheter images are shown in Fig.\,\ref{fig:correct}. In these cases, multiple catheters appear to be detected in their intuitive regions (near the neck for NGT and ETT, and near the pelvis for umbilical catheters). Examples for unsuccessful predictions are shown in Fig.\,\ref{fig:incorrect}, one for each catheter type. These images can be a useful proxy for how well the network is training; that is, are the areas of greatest influence where one would expect them to be? The NGT is clearly visible though the prediction remained slightly below our cut-off, perhaps due to some similarity with ECG lines (which the network has been trained to generally ignore). In any case the abdomen remains the region of primary influence. The ETT LAM shows a plausible region of influence yet the network fails to correctly predict this image; this ETT is placed somewhat shorter than average and crosses at its tip with the NGT. UVCs and UACs may be confused with each other, or malpositioned one for the other, and this may be the case for the two examples shown.

\section*{Discussion}
Our multi-label CNN demonstrates strong performance in the simultaneous detection of up to four catheters of interest. This result was robust to the presence of single or multiple catheters. While it may be challenging to discern the full breadth of possibilities for reasons of success or failure in label prediction of individual cases, there are some noteworthy features of the relatively common catheters we have chosen in this work which might lead to some insight. These may be considered on a spectrum from low-level features, such as edges, lines, and pixel intensities, to high-level features such as the relationship between the catheter to other objects in the image. First, there is a characteristic anatomic location that each catheter is expected to occupy, although for some this may be quite broad. Second, the course of the catheter relative to itself or to certain landmarks, such as ribs or vertebrae, is important. These two features provide information that may be useful both for identification and analysis of correctness of placement. Third, there is the intensity profile of the catheter’s tubing. In addition to these intrinsic features, there may also be useful information provided by the overall radiograph. For example, in some cases the existence of other catheters, or the presence of pathology, might increase the likelihood of the presence of additional catheters (e.g., sicker patients may require more catheters). Our evaluation metrics and LAMs provide useful clues as to the features the network finds most important for prediction.

\begin{SCfigure*}[\sidecaptionrelwidth][ht]
\centering
\includegraphics[width=12.4cm]{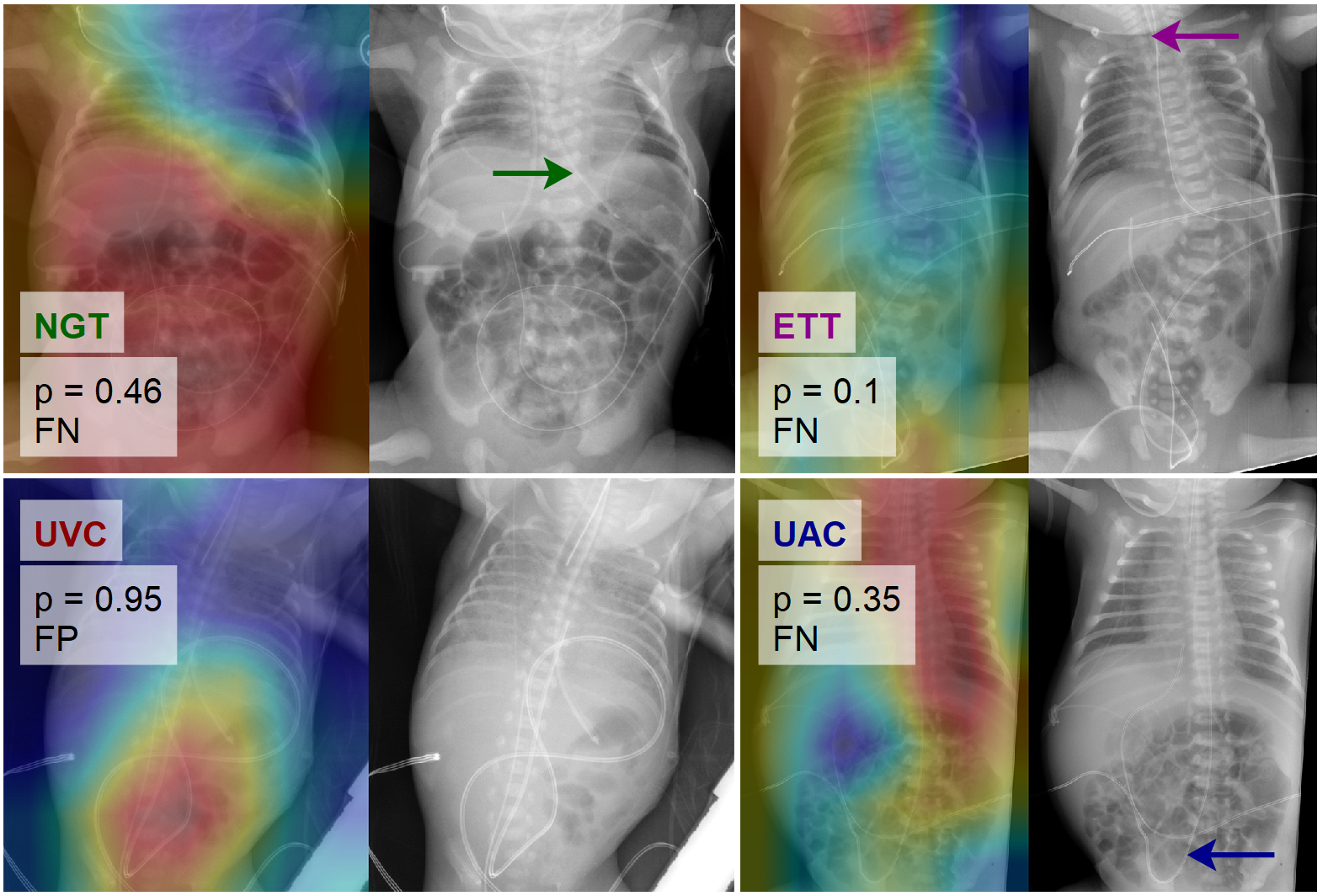}
\caption{Examples of CAMs for unsuccessful catheter classification. Arrows indicate catheters present but not detected according to our chosen probability cut-off (FNs). See Fig. 2 and the text for additional discussion.}
\label{fig:incorrect}
\end{SCfigure*}

Considering other studies and the physical features of our catheters, we first compare the performance of NGT and ETT prediction. These catheters originate close together on the images, but appear very different in length and intensity profile. Template matching was employed by Ramakrishna et al. \cite{Ramakrishna2011}, in which NGTs and ETTs were distinguished by their intensity profiles with sensitivities of 76.5\% and 73.7\%, and specificities of 84.0\% and 91.3\%, respectively. Using anatomic arguments for where catheters are generally expected, Sheng, Li and Pei \cite{Sheng2009} detect the presence of ETTs with 94\% sensitivity and NGTs with 82\% sensitivity.  Other studies have used support vector machines to yield sensitivities in region of 94-95\% \cite{Chen2016,Kao2015}. In comparison with our results, we achieve a very high overall AP for NGTs and ETTs (over 97\% each), and high sensitivity of 98\% for NGTs, however several false negatives (8 of 47 cases) decreases our sensitivity to 83\% for ETTs. Interestingly, ETT sensitivity is the only case where the single-label network (98\%) outperforms the multi-label network. This is despite the LAMs indicating the appropriate region of ETTs is examined by the multi-label network even in incorrect cases (see Fig.\,\ref{fig:incorrect}, top right). It is possible that the potential existence of another tube in this same region (an NGT) leads to confusion in the multi-label network. 

Examining umbilical catheters (UACs and UVCs), these enter the abdomen close together (through the umbilicus) but differ in their course relative to the spine and cardiac silhouette. As well, unlike the difference in appearance of ETTs and NGTs, UACs and UVCs are identical in physical structure. This removes the possibility of intensity information contributing to their classification. Importantly, a UAC will often have a downward course as it enters the aorta via the iliac arteries, unlike a UVC which progresses upward and somewhat left (patient right) as it enters the inferior vena cava. Normal anatomic variants and common misplacement can lead to uncertainty for both the network and radiologists. For example, UVCs can be mistakenly placed in the portal vein, though it is much less common for UACs to take a lateral deviation as the catheter is advanced \cite{Concepcion2017,Ramasethu2008}. No other studies of which we are aware have specifically examined UAC and UVC classification, although Yi et al. \cite{Yi2019} were able to detect both types of umbilical catheters in a network which was not specific to catheter type and detected only pixel locations of catheters of interest. The differences and challenges outlined above for the classification of UACs and UVCs appear to manifest in our network: a reduced overall AP and specificity for UVCs (0.94 and 79\%, respectively) compared to UACs (0.98 and 100\%, respectively). The greater number of false positive UVCs may also be due, in part, to portions of the catheter which are external to the patient misleadingly projecting over the abdomen so that it appears to be following an arterial course, in which case there may be a “false” trajectory confusing the network.

As discussed in the introduction, a broader consideration is the performance of our network in situations where there are multiple catheters. Apart from the studies previously mentioned, only one other has explored the consideration of multiple catheters: Subramanian et al. \cite{Subramanian2019} classified multiple types of central venous catheters with a random forest model, and achieved a precision of 95\%. Analogous work has been done on other non-catheter radiographic findings, notably on chest X-rays (CXRs). For instance, Wang et al. also used ResNet-50 in a multi-label setup to identify 8 possible pathologies on adult CXRs, and achieved AP values ranging from 0.56 – 0.81 \cite{Wang2019}. Interestingly, our overall network performance in terms of AP did not decrease for the subset of data containing two or more catheters (maintaining a range of 0.94 - 1.00). Further, for most of our raw measures the multi-label network outperforms the catheter-specific single-label networks, confirming that there is at least some useful contribution of information pertaining to the presence or absence of other catheters. This provides confidence that our network is robust to the presence of multiple catheters, a situation commonly encountered in neonatal cases where breathing, nutrition, monitoring and medication support often coincide.

With respect to our choice of network, the current approach was based on ResNet-50. While other more advanced network architectures such as ResNeXt \cite{Xie2017} and DenseNet \cite{Huang2017} might further improve performance, we found ResNet-50 to be a reasonable compromise between model size and performance. Segmentation methods that may isolate catheters from the image could also be incorporated to further improve the performance. However, given the current classification performance of our model, we argue that adding another layer of complexity is unnecessary. Further, we highlight that the proposed method is a critical component that builds toward a fully automatic catheter evaluation system.

Two noteworthy limitations of the above work pertain to our sample size and generalizability. In brief, labelling a large sample of images manually is a highly accurate but time-consuming exercise. Whilst we have attained a high level of model performance, our confidence intervals are somewhat broadened from the use of only 78 test images. It is also unclear how this network might perform with external validation on data gathered from another institution, as this was a single-site study, or on adult chest radiographs (which would not contain UACs or UVCs); such exploration has been set for future work, along with the two remaining questions on catheter detection (tip location and appropriate placement). The predictions of our network can nevertheless be useful for prepopulating radiology reports, although additional work is required to cast this tool into an efficient and effective component of the radiologist’s workflow. 

\section*{Conclusions}
We have constructed a multi-label CNN that accurately detects the presence or absence of four catheters of interest (NGTs, ETTs, UACs, UVCs) on neonatal chest/abdominal radiographs. The presence of multiple catheters yielded results that were consistent with the overall AP measurements. Our multi-label approach provides for a model that is more efficient and less complex than individual networks for each catheter of interest. In sum, we have a robust network which answers the question “what types of catheters are present?”, which contributes to our overall goal of constructing a tool that automatically detects correct placement of various catheters in radiographs. Such a tool could assist the workflow of radiologists by flagging cases for urgent assessment or augmenting diagnostic imaging reports. Our next steps are to generate analogous networks to answer the remaining questions for catheter placement: tip location and correct placement. These will be featured in future work.

\section*{Acknowledgements}
RDEH gratefully acknowledges helpful discussions with\linebreak Dr.\  Carolyn Augusta.

\section*{References}

\renewcommand\thefigure{S\arabic{figure}}
\setcounter{figure}{0}
\renewcommand\thetable{S\arabic{table}}
\setcounter{table}{0}
\section*{SUPPLEMENTARY INFORMATION}
The figure and tables shown on the following pages are supplied as supplementary information, as referenced in the main text.

\begin{figure*}[tbhp]
\centering
\includegraphics[width=6in]{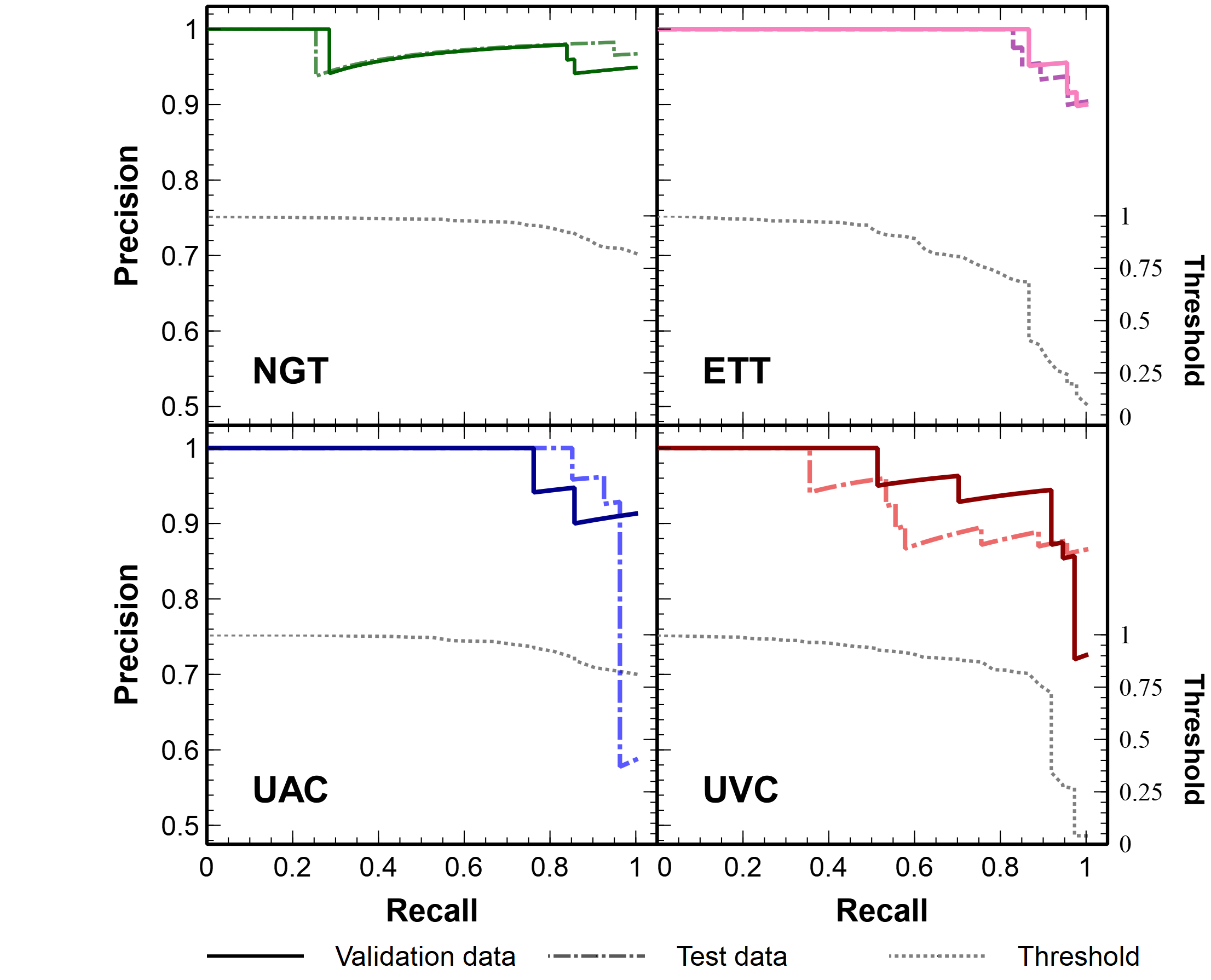}
\caption{Precision-Recall curves (from the validation dataset) for our 4 catheters of interest. We show results for the entire validation dataset (solid curves) and for the test images (dash-dot curves). For each catheter we also plot the accompanying thresholds for each recall value (dotted grey curves). NGT – nasogastric tube; ETT – endotracheal tube; UAC – umbilical arterial catheter; UVC – umbilical venous catheter.}
\label{fig:prvalsupp}
\end{figure*}

\begin{table*}[ht]
\centering
\caption{Detailed results for each catheter type from our model on 78 test cases, split by number of labels per image; an expanded version of Table 2 of the main text.}
\begin{tabular}{llllll}
\# Labels        & NGT                     & ETT                     & UAC                     & UVC                     & All$^f$                    \\
\midrule
                 & \multicolumn{5}{c}{Average Precision}                                                                                           \\ 
\cmidrule(lr){2-6}
0$^a$               &--                      &--                      &--                      &--                      &--                      \\
1$^b$               & 1 ( - )                 &--                      &--                      & 1 ( - )                 &--                      \\
2                & 1 ( - )                 & 0.991 (0.089 –   1.000) & 0.850 (0.624 –   0.951) & 0.935 (0.632 –   0.992) &--                      \\
3                & 0.947 (0.036 –   1.000) & 1 ( - )                 & 0.971 (0.500 –   0.999) & 0.884 (0.263 –   0.994) &--                      \\
4$^c$               &--                      & 1 ( - )                 & 1 ( - )                 &--                      &--                      \\
\textgreater{}1d & 0.975 (0.255 –   1.000) & 0.997 (0.009 –   1.000) & 0.981 (0.797   - 0.998) & 0.937 (0.689 –   0.990) &--                      \\
0-4              & 0.977 (0.679 –   0.999) & 0.989 (0.751 –   1.000) & 0.979 (0.873   - 0.997) & 0.937 (0.785 –   0.984) &--                      \\
Single$^e$          & 0.976 (0.683 –   0.999) & 0.982 (0.794 –   0.999) & 0.979 (0.873   - 0.997) & 0.917 (0.763 –   0.975) &--                      \\
\cmidrule(lr){2-6}
                 & \multicolumn{5}{c}{AUROC}                                                                                                       \\
\cmidrule(lr){2-6}
0$^a$               &--                      &--                      &--                      &--                      &--                      \\
1$^b$               & 1 ( - )                 &--                      &--                      & 1 ( - )                 &--                      \\
2                & 1 ( - )                 & 0.985 (0.502 – 1.000)   & 0.925 (0.230 – 0.998)   & 0.957 (0.512 – 0.998)   &--                      \\
3                & 0.677 (0.431 – 0.852)   & 1 ( - )                 & 0.977 (0.291 – 1.000)   & 0.633 (0.377 – 0.832)   &--                      \\
4$^c$               &--                      &--                      &--                      &--                      &--                      \\
\textgreater{}1d & 0.894 (0.775 – 0.953)   & 0.987 (0.860 – 0.999)   & 0.976 (0.776 – 0.998)   & 0.886 (0.746 – 0.954)   &--                      \\
0-4              & 0.958 (0.865 – 0.988)   & 0.984 (0.863 – 0.998)   & 0.983 (0.756 – 0.999)   & 0.928 (0.806 – 0.976)   &--                      \\
Single$^e$          & 0.948 (0.853 – 0.983)   & 0.969 (0.857 – 0.994)   & 0.986 (0.738 – 0.999)   & 0.896 (0.768 – 0.958)   &--                      \\
\cmidrule(lr){2-6}
                 & \multicolumn{5}{c}{Sensitivity}                                                                                                 \\
\cmidrule(lr){2-6}
0$^a$               &--                      &--                      &--                      &--                      &--                      \\
1$^b$               & 1 ( - )                 &--                      &--                      & 1 ( - )                 & 1 ( - )                 \\
2                & 0.947   (0.706 - 0.993) & 0.867   (0.595 - 0.966) & 0.500   (0.123 - 0.877) & 1 (-)                   & 0.896   (0.773 - 0.956) \\
3                & 0.941   (0.680 - 0.992) & 0.941   (0.680 - 0.992) & 0.625   (0.285 - 0.875) & 0.933   (0.648 - 0.991) & 0.895   (0.785 - 0.952) \\
4$^c$               & 0.933   (0.648 - 0.991) & 0.933   (0.648 - 0.991) & 0.867   (0.595 - 0.966) & 0.933   (0.648 - 0.991) & 0.917   (0.815 - 0.965) \\
\textgreater{}1d & 0.941   (0.833 - 0.981) & 0.915   (0.794 - 0.968) & 0.741   (0.547 - 0.871) & 0.950   (0.821 - 0.987) & 0.903   (0.848 - 0.940) \\
0-4              & 0.949   (0.854 - 0.984) & 0.915   (0.794 - 0.968) & 0.741   (0.547 - 0.871) & 0.956   (0.839 - 0.989) & 0.910   (0.858 - 0.944) \\
Single$^e$          & 0.947   (0.706 - 0.993) & 0.867   (0.595 - 0.966) & 0.500   (0.123 - 0.877) & 1 (-)                   & 0.896   (0.773 - 0.956) \\
\cmidrule(lr){2-6}
                 & \multicolumn{5}{c}{Specificity}                                                                                                 \\
\cmidrule(lr){2-6}
0$^a$               & 0.857 (0.419 – 0.980)   & 1 ( - )                 & 1 ( - )                 & 0.857 (0.419 – 0.980)   & 0.929 (0.755 – 0.982)   \\
1$^b$               & 1 ( - )                 & 0.923 (0.609 – 0.989)   & 1 ( - )                 & 1 ( - )                 & 0.974 (0.839 – 0.996)   \\
2                & 1 ( - )                 & 1 ( - )                 & 1 ( - )                 & 0.857 (0.573 – 0.964)   & 0.958 (0.848 – 0.990)   \\
3                & 0.500 (0.059 – 0.941)   & 1 ( - )                 & 1 ( - )                 & 0 ( - )                 & 0.737 (0.502 – 0.886)   \\
4$^c$               &--                      &--                      &--                      &--                      &--                      \\
\textgreater{}1d & 0.857 (0.419 – 0.980)   & 1 ( - )                 & 1 ( - )                 & 0.667 (0.429 – 0.842)   & 0.896 (0.797 – 0.949)   \\
0-4              & 0.895 (0.663 – 0.974)   & 0.968 (0.804 – 0.995)   & 1 ( - )                 & 0.788 (0.617 – 0.895)   & 0.925 (0.867 – 0.959)   \\
Single$^e$          & 0.842 (0.608 – 0.948)   & 0.710 (0.530 – 0.841)   & 0.980 (0.874 – 0.997)   & 0.788 (0.617 – 0.895)   & 0.851 (0.780 – 0.902)   \\
\cmidrule(lr){2-6}
                 & \multicolumn{5}{c}{Hamming Lossg}                                                                                               \\
\cmidrule(lr){2-6}
0$^a$               & 0.143 (0.02 – 0.581)    & 0 ( - )                 & 0 ( - )                 & 0.143 (0.02 – 0.581)    & 0.071 (0.018 – 0.245)   \\
1$^b$               & 0 ( - )                 & 0.077 (0.011 – 0.391)   & 0 ( - )                 & 0 ( - )                 & 0.019 (0.003 – 0.124)   \\
2                & 0.042 (0.006 – 0.244)   & 0.083 (0.021 – 0.279)   & 0.042 (0.006 – 0.244)   & 0.083 (0.021 – 0.279)   & 0.063 (0.028 – 0.132)   \\
3                & 0.053 (0.007 – 0.294)   & 0.158 (0.052 – 0.392)   & 0.105 (0.026 – 0.337)   & 0.211 (0.081 – 0.446)   & 0.132 (0.072 – 0.228)   \\
4$^c$               & 0 ( - )                 & 0.200 (0.066 – 0.470)   & 0.067 (0.009 – 0.352)   & 0 ( - )                 & 0.067 (0.025 – 0.165)   \\
\textgreater{}1d & 0.034 (0.009 – 0.128)   & 0.138 (0.071 – 0.252)   & 0.069 (0.026 – 0.170)   & 0.103 (0.047 – 0.212)   & 0.086 (0.056 – 0.130)   \\
0-4              & 0.038 (0.012 – 0.113)   & 0.115 (0.061 – 0.207)   & 0.051 (0.019 – 0.129)   & 0.090 (0.043 – 0.176)   & 0.074 (0.049 – 0.108)   \\
Single$^e$          & 0.051 (0.019 – 0.129)   & 0.128 (0.070 – 0.222)   & 0.064 (0.027 – 0.145)   & 0.141 (0.080 – 0.237)   & 0.096 (0.068 – 0.134)  \\ \bottomrule
\end{tabular}
\label{tab:resultssupp}
\addtabletext{$~$\\
Confidence intervals are set at 95\%, computed as described in the main text. Values of 0 or 1 have undefined confidence intervals ( - ).\\
NGT – nasogastric tube; ETT – endotracheal tube; UAC – umbilical arterial catheter; UVC – umbilical venous catheter; AUROC – area under the receiver operating characteristic curve.\\
$^a$ Dashes in these rows indicate that: AP cannot be computed due to the absence of true positive cases (precision = 0 for all recall); similarly AUROC cannot be computed; sensitivity is undefined here.\\
$^b$ Dashes in this column indicate that AP cannot be computed due to the absence of both true positive and false negative cases. Recall (sensitivity) is thus undefined in such cases.\\
$^c$ Dashes in this column indicate that AP cannot be computed due to the absence of both false positive and false negative cases (precision = 1 and recall = 1, always); specificity is undefined here.\\
$^d$ Combined results for those images containing multiple (2 or more) catheters.\\
$^e$ Single-label results are for individual binary classification networks for each type of catheter.\\
$^f$ Combined results for all catheter types. When computing these measures for all catheter types, there is some dependence among the data points since most of the images contain multiple catheters which may influence each other. Thus, this metric is not considered robust.\\
$^g$ Note that, unlike the three previous measures, values closer to zero represent better performance. See the text for details.\\}
\end{table*}

\begin{table*}[ht]
\centering
\caption{Details of the test dataset results on 78 test cases by catheter type and number of labels, for both the multi-label and single-label networks.}
\begin{tabular}{llcccccccc}
 &  & \multicolumn{7}{l}{Multi-label network} & Single-label network \\
 &  & \multicolumn{7}{l}{Number of labels per image:} &  \\
\cmidrule(lr){3-9} \cmidrule(lr){10-10}
 &  & 0 & 1 & 2 & 3 & 4 & \textgreater{}1 & 0 – 4 & 0 - 4 \\ \midrule
\multirow{4}{*}{NGT} & TN & 6 & 5 & 5 & 1 & 0 & 6 & 17 & 16 \\
 & FP & 1 & 0 & 0 & 1 & 0 & 1 & 2 & 3 \\
 & FN & 0 & 0 & 1 & 1 & 1 & 3 & 3 & 1 \\
 & TP & 0 & 8 & 18 & 16 & 14 & 48 & 56 & 58 \\ \midrule
\multirow{4}{*}{ETT} & TN & 7 & 11 & 8 & 2 & 0 & 10 & 28 & 22 \\
 & FP & 0 & 2 & 1 & 0 & 0 & 1 & 3 & 9 \\
 & FN & 0 & 0 & 2 & 1 & 1 & 4 & 4 & 1 \\
 & TP & 0 & 0 & 13 & 16 & 14 & 43 & 43 & 46 \\ \midrule
\multirow{4}{*}{UAC} & TN & 7 & 13 & 20 & 11 & 0 & 31 & 51 & 50 \\
 & FP & 0 & 0 & 0 & 0 & 0 & 0 & 0 & 1 \\
 & FN & 0 & 0 & 2 & 3 & 2 & 7 & 7 & 4 \\
 & TP & 0 & 0 & 2 & 5 & 13 & 20 & 20 & 23 \\ \midrule
\multirow{4}{*}{UVC} & TN & 6 & 8 & 12 & 1 & 0 & 13 & 26 & 26 \\
 & FP & 1 & 0 & 2 & 3 & 0 & 5 & 7 & 7 \\
 & FN & 0 & 0 & 0 & 1 & 1 & 2 & 0 & 4 \\
 & TP & 0 & 5 & 10 & 14 & 14 & 38 & 45 & 41 \\ \bottomrule
\end{tabular}
\label{tab:counts}
\addtabletext{$~$\\
Note: layout is analogous to Table 2 from the main text. TN – true negatives; FP – false positives; FN – false negatives; TP – true positives; NGT – nasogastric tube; ETT – endotracheal tube; UAC – umbilical arterial catheter; UVC – umbilical venous catheter; AUROC – area under the receiver operating characteristic curve.\\}
\end{table*}


\begin{thebibliography}{10}

\bibitem{Schmidt2008}
Schmidt U, et~al. (2008) {Tracheostomy tube malposition in patients admitted to
  a respiratory acute care unit following prolonged ventilation}.
\newblock {\em Chest} 134(2):288--294.

\bibitem{Remerand2007}
Rem{\'{e}}rand F, et~al. (2007) {Incidence of chest tube malposition in the
  critically ill: A prospective computed tomography study}.
\newblock {\em Anesthesiology} 106(6):1112--1119.

\bibitem{Thomas1998}
Thomas BW, Falcone RE (1998) {Confirmation of nasogastric tube placement by
  colorimetric indicator detection of carbon dioxide: a preliminary report}.
\newblock {\em Journal of the American College of Nutrition} 17(2):195--197.

\bibitem{Godoy2012}
Godoy MCB, Leitman BS, {De Groot} PM, Vlahos I, Naidich DP (2012) {Chest
  radiography in the ICU: Part 1, evaluation of airway, enteric, and pleural
  tubes}.

\bibitem{MacDonald2013}
MacDonald MG, Ramasethu J, Rais-Bahrami K (2013) {\em {Atlas of Procedures in
  Neonatology}}.
\newblock (Lippincott Williams {\&} Wilkins, Philadelphia), 5th ed edition.

\bibitem{Green1998}
Green C, Yohannan MD (1998) {Umbilical arterial and venous catheters:
  placement, use, and complications.}
\newblock {\em Neonatal network : NN} 17(6):23--8.

\bibitem{Concepcion2017}
Concepcion NDP, Laya BF, Lee EY (2017) {Current updates in catheters, tubes and
  drains in the pediatric chest: A practical evaluation approach}.

\bibitem{Keller2007a}
Keller BM, Reeves AP, Cham MD, Henschke CI, Yankelevitz DF (2007)
  {Semi-automated location identification of catheters in digital chest
  radiographs} in {\em Medical Imaging 2007: Computer-Aided Diagnosis}.
\newblock Vol.{} 6514, p. 65141O.

\bibitem{Huo2007}
Huo Z, Chen S, Foos D, Rao Y (2007) {Computer-aided detection of tubes and
  lines in portable chest X-ray images}.
\newblock {\em International Journal of Computer Assisted Radiology and
  Surgery} 2(Suppl. 1):S370--S372.

\bibitem{Brunelli2009a}
Brunelli R (2009) {\em {Template Matching Techniques in Computer Vision: Theory
  and Practice}}.
\newblock pp. 1--338.

\bibitem{Duda1972}
Duda RO, Hart PE (1972) {Use of the Hough transformation to detect lines and
  curves in pictures}.
\newblock {\em Communications of the ACM} 15(1):11--15.

\bibitem{Lakhani2017}
Lakhani P (2017) {Deep Convolutional Neural Networks for Endotracheal Tube
  Position and X-ray Image Classification: Challenges and Opportunities}.
\newblock {\em Journal of Digital Imaging} 30(4):460--468.

\bibitem{Singh2019}
Singh V, Danda V, Gorniak R, Flanders A, Lakhani P (2019) {Assessment of
  Critical Feeding Tube Malpositions on Radiographs Using Deep Learning}.
\newblock {\em Journal of Digital Imaging} 32:651--655.

\bibitem{Mercan2014}
Mercan CA, Celebi MS (2014) {An approach for chest tube detection in chest
  radiographs}.
\newblock {\em IET Image Processing} 8(2):122--129.

\bibitem{Lee2018}
Lee H, Mansouri M, Tajmir S, Lev MH, Do S (2018) {A Deep-Learning System for
  Fully-Automated Peripherally Inserted Central Catheter (PICC) Tip Detection}.
\newblock {\em Journal of Digital Imaging} 31(4):393--402.

\bibitem{Yi2019}
Yi X, Adams S, Babyn P, Elnajmi A (2019) {Automatic catheter and tube detection
  in pediatric X-ray images using a scale-recurrent network and synthetic
  data}.
\newblock {\em Journal of Digital Imaging}.

\bibitem{Frid-Adar2019}
Frid-Adar M, Amer R, Greenspan H (2019) {Endotracheal Tube Detection and
  Segmentation in Chest Radiographs Using Synthetic Data} in {\em Lecture Notes
  in Computer Science (including subseries Lecture Notes in Artificial
  Intelligence and Lecture Notes in Bioinformatics)}.
\newblock Vol.{} 11769 LNCS, pp. 784--792.

\bibitem{Subramanian2019}
Subramanian V, et~al. (2019) {Automated Detection and Type Classification of
  Central Venous Catheters in Chest X-Rays} in {\em Lecture Notes in Computer
  Science (including subseries Lecture Notes in Artificial Intelligence and
  Lecture Notes in Bioinformatics)}.
\newblock Vol.{} 11769 LNCS, pp. 522--530.

\bibitem{Yi2020a}
Yi X, Adams SJ, Henderson RDE, Babyn P (2020) {Computer-aided Assessment of
  Catheters and Tubes on Radiographs: How Good Is Artificial Intelligence for
  Assessment?}
\newblock {\em Radiology: Artificial Intelligence} 2(1):e190082.

\bibitem{Sheng2009}
Sheng C, Li L, Pei W (2009) {Automatic detection of supporting device
  positioning in intensive care unit radiography}.
\newblock {\em The International Journal of Medical Robotics and Computer
  Assisted Surgery} 5(3):332--340.

\bibitem{Ramakrishna2011}
Ramakrishna B, Brown M, Goldin J, Cagnon C, Enzmann D (2011) {Catheter
  detection and classification on chest radiographs: an automated prototype
  computer-aided detection (CAD) system for radiologists}, eds.{} Summers RM,
  van Ginneken B.
\newblock (International Society for Optics and Photonics), Vol.{} 7963, p.
  796333.

\bibitem{Chen2016}
Chen S, Zhang M, Yao L, Xu W (2016) {Endotracheal tubes positioning detection
  in adult portable chest radiography for intensive care unit}.
\newblock {\em International Journal of Computer Assisted Radiology and
  Surgery} 11(11):2049--2057.

\bibitem{Kao2015}
Kao EF, Jaw TS, Li CW, Chou MC, Liu GC (2015) {Automated detection of
  endotracheal tubes in paediatric chest radiographs}.
\newblock {\em Computer Methods and Programs in Biomedicine} 118(1):1--10.

\bibitem{He2016a}
He K, Zhang X, Ren S, Sun J (2016) {Deep residual learning for image
  recognition} in {\em Proceedings of the IEEE Computer Society Conference on
  Computer Vision and Pattern Recognition}.
\newblock Vol.{} 2016-Decem, pp. 770--778.

\bibitem{Tsoumakas2007}
Tsoumakas G, Katakis I (2007) {Multi-label classification: An overview} in {\em
  International Journal of Data Warehousing and Mining}.
\newblock Vol.{}~3, pp. 1--13.

\bibitem{Zou2013}
Zou KH, Yu CR, Liu K, Carlsson MO, Cabrera J (2013) {Optimal thresholds by
  maximizing or minimizing various metrics via ROC-type analysis}.
\newblock {\em Academic Radiology} 20(7):807--815.

\bibitem{Zhou2016}
Zhou B, Khosla A, Lapedriza A, Oliva A, Torralba A (2016) {Learning Deep
  Features for Discriminative Localization} in {\em Proceedings of the IEEE
  Computer Society Conference on Computer Vision and Pattern Recognition}.
\newblock Vol.{} 2016-Decem, pp. 2921--2929.

\bibitem{Sternberg2001}
Sternberg MR, Hadgu A (2001) {A GEE approach to estimating sensitivity and
  specificity and coverage properties of the confidence intervals}.
\newblock {\em Statistics in Medicine} 20(9-10):1529--1539.

\bibitem{Boyd2013}
Boyd K, Eng KH, Page CD (2013) {Area under the precision-recall curve: Point
  estimates and confidence intervals} in {\em Lecture Notes in Computer Science
  (including subseries Lecture Notes in Artificial Intelligence and Lecture
  Notes in Bioinformatics)}.
\newblock Vol.{} 8190 LNAI, pp. 451--466.

\bibitem{Ramasethu2008}
Ramasethu J (2008) {Complications of Vascular Catheters in the Neonatal
  Intensive Care Unit}.

\bibitem{Wang2019}
Wang X, et~al. (2019) {ChestX-ray: Hospital-Scale Chest X-ray Database and
  Benchmarks on Weakly Supervised Classification and Localization of Common
  Thorax Diseases} in {\em Advances in Computer Vision and Pattern
  Recognition}.
\newblock pp. 369--392.

\bibitem{Xie2017}
Xie S, Girshick R, Doll{\'{a}}r P, Tu Z, He K (2017) {Aggregated residual
  transformations for deep neural networks} in {\em Proceedings - 30th IEEE
  Conference on Computer Vision and Pattern Recognition, CVPR 2017}.
\newblock Vol.{} 2017-Janua, pp. 5987--5995.

\bibitem{Huang2017}
Huang G, Liu Z, {Van Der Maaten} L, Weinberger KQ (2017) {Densely connected
  convolutional networks} in {\em Proceedings - 30th IEEE Conference on
  Computer Vision and Pattern Recognition, CVPR 2017}.
\newblock Vol.{} 2017-Janua, pp. 2261--2269.

\end{thebibliography}
\end{document}